%% file: perturber_demo.tex
\pdfoutput=1
\documentclass[11pt]{article}

\usepackage[preprint]{acl}

\usepackage{times}
\usepackage{latexsym}
\usepackage[T1]{fontenc}
\usepackage[utf8]{inputenc}
\usepackage{microtype}
\usepackage{inconsolata}
\usepackage{graphicx}
\usepackage{booktabs}
\usepackage{tabularx}
\usepackage{tikz}
\usetikzlibrary{arrows.meta, positioning, fit, backgrounds}
\usepackage{url}

\graphicspath{{figures/}}

\title{perturber: A Stateless Service for Semantics-Preserving Text Perturbations}

\author{Khoi-Nguyen Tran \\
  IBM Research \\
  \texttt{kndtran@ibm.com}
}

\begin{document}
\maketitle

\begin{abstract}

Large language model (LLM) evaluations on clean benchmarks overstate model robustness: real
inputs arrive perturbed, and these meaning-preserving variants routinely move model answers.
Studying this systematically requires applying perturbations one at a time, reproducibly, and
identically wherever they run, yet existing tools couple them to a benchmarking stack rather
than packaging them as a reusable transform layer. We present \texttt{perturber}, a stateless
service and Python library that applies exactly one semantics-preserving perturbation per call,
drawn from a first-party catalog of 68 perturbations across 11 families. It turns on a single primitive,
with composition left to the caller chaining calls, so the service and the library produce
identical output by construction. We evaluate the tool itself, its correctness,
reproducibility, and cost, rather than any model's robustness, measuring meaning-preservation
per perturbation rather than assuming it. The demonstration adds a live playground, an
introspectable catalog for benchmark selection, and a one-command reproducible evaluation.

\end{abstract}

\section{Introduction}
\label{sec:intro}

Large language models (LLMs) are evaluated on clean, static benchmarks, but they are used on
messy text: real inputs arrive with typos, inconsistent casing, ad hoc formatting, and
reworded phrasing. These changes leave the meaning intact, yet a growing body of work shows
that they move model answers. A recent study reports accuracy degradation of up to 12\%, a
reversal of the ranking between two models under a single perturbation in 63\% of cases, and
input phrasing accounting for as much as half of a model's measured performance variance
\citep{romanou2026brittlebench}. A benchmark run on clean text can therefore overstate a
model's robustness, and can even rank a more fragile model above a more stable one.

Measuring this fragility credibly imposes strict demands on the perturbation step: each
transformation must be applied one at a time and reproducibly, its provenance must be clear, and
it must behave identically whether a harness calls it as a library, from a shell, or over the
network. These are demands on tooling, and no existing tool meets them as a standalone transform.
BrittleBench, for instance, bundles the perturbations with model inference and scoring rather
than exposing a transform layer a harness can call in isolation. So each study reimplements the
perturbations it needs, and small differences between those reimplementations leave robustness
numbers from different studies not directly comparable.

We present \texttt{perturber}, a stateless service and Python library that fills this gap. It
serves no model, computes no score, and holds no state; it is the transformation layer a
robustness harness invokes, exposing a first-party catalog of 68 perturbations across 11 families. Our
contributions are fourfold. First, and most important, a standard set of perturbation
\emph{definitions}: each of the 68 is named, scoped, and typed once, so two studies applying the
same perturbation apply the same transformation and their robustness numbers become comparable.
Second, a single-primitive design, \texttt{perturb(text, name, params, seed)}, in which
composition is the caller chaining calls, so the library and the network service are byte-identical
by construction and the deterministic tier is a pure function of its arguments, a reproducibility
contract augmentation libraries do not offer. Third, an introspectable catalog from which a
downstream harness selects its own benchmark, with an optional tier that wraps an external engine
for direct comparison. Fourth, a reproducible fidelity harness that measures
semantics-preservation per perturbation rather than assuming it, catching two defects in the
transform layer. \texttt{perturber} is open source and pip-installable at
\url{https://github.com/IBM/perturber}.

\section{System Overview}
\label{sec:system}

\begin{figure}[t]
  \centering
  \begin{tikzpicture}[
      font=\small,
      >={Stealth[length=2mm]},
      iface/.style={draw, rounded corners=2pt, minimum width=1.9cm, minimum height=0.7cm, align=center, fill=black!3},
      core/.style={draw, rounded corners=2pt, minimum width=4.6cm, minimum height=0.8cm, align=center, fill=black!5},
      reg/.style={draw, rounded corners=2pt, minimum width=4.6cm, minimum height=0.8cm, align=center, fill=black!5},
      node distance=0.55cm and 0.45cm,
    ]
    \node[iface] (lib) {Library\\\texttt{perturb()}};
    \node[iface, right=of lib] (cli) {CLI\\\texttt{perturber}};
    \node[iface, right=of cli] (http) {Service\\\texttt{POST /perturb}};
    \node[core, below=of cli] (prim) {One primitive\\\texttt{perturb(text, name, params, seed)}};
    \node[reg, below=of prim] (registry) {Registry\\\footnotesize category, family, typed params, seed-sensitivity};
    \draw[->] (lib.south) -- (prim.north west);
    \draw[->] (cli.south) -- (prim.north);
    \draw[->] (http.south) -- (prim.north east);
    \draw[<->] (prim.south) -- node[right, font=\footnotesize] {resolves} (registry.north);
  \end{tikzpicture}
  \caption{The architecture. The library, the CLI, and the network service are three thin
  interfaces over one primitive, which resolves every perturbation through one registry. Because
  the interfaces share this path, a chain produces byte-identical output whether it is run in the
  library or over HTTP.}
  \label{fig:arch}
\end{figure}
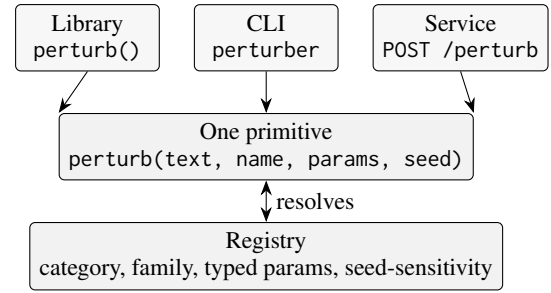

The system is built on one primitive,
\texttt{perturb(text, name, params, seed, category)}, which applies exactly one perturbation
and returns the transformed string (Figure~\ref{fig:arch}). There is deliberately no
composition grammar and no stack; a caller composes perturbations by chaining calls, feeding
the output of one into the next. For the deterministic tier the primitive is a pure function of
\texttt{(text, name, params, seed)}: it reseeds the pseudo-random generators under a lock before
each call, defaults the seed to zero, and admits no wall-clock or process randomness, so a
repeated call reproduces its output exactly. The model-backed tier is best-effort and
non-deterministic by nature, and we exclude it from every determinism and latency claim below.

Three layers sit around this primitive. The \emph{library} exposes \texttt{perturb} in process,
for a Python harness that wants the transform directly. The \emph{command-line tool} wraps it for
the shell, reading from an argument or standard input and writing the result verbatim to standard
output, which suits quick inspection and scripting. The \emph{service} exposes it over HTTP, with
one typed route per perturbation so a harness in any language can validate its parameters and drive
the catalog remotely. Because every layer forwards to the same primitive, a chain produces
byte-identical output whichever a caller uses. In practice a harness reads the catalog, filters it
on whatever field it cares about, and applies each perturbation one at a time.

A single registry binds the layers together, the abstraction boundary of the system:
everything above it sees one uniform primitive, and everything below it is a perturbation
definition tagged with its provenance and scope. Every perturbation registers its category,
family, typed parameters with defaults and permitted choices, seed-sensitivity, and an optional
system prompt, and all three layers, with the typed HTTP routes, are generated from that one
description.

A harness can thus enumerate the catalog and select its own subset rather than work
against a hard-coded list. Because a perturbation is just a registry entry, an
external perturbation design enters here as a new provenance-tagged source without changing the
primitive or the interfaces above it; the \texttt{brittlebench} compatibility tier
(Section~\ref{sec:catalog}) is exactly such a source. The system is open source and installable
with \texttt{pip}; the service starts with one command and ships as a container image that runs
unchanged anywhere. Holding no state, it carries no configuration or data-migration burden.

\section{The Perturbation Catalog}
\label{sec:catalog}

The first-party catalog comprises 68 perturbations across 11 families.\footnote{The public
release ships these 68. One further first-party perturbation, \texttt{persona\_instruction},
is described in the appendix but withheld from the public release because it draws on PersonaHub
\citep{ge2024personahub}, which is licensed CC-BY-NC-SA-4.0 for research use only
(see the Ethics Statement).} Every perturbation carries a
mechanism: it changes the surface form a human reads as unchanged, yet moves the token
sequence the model consumes. This is the common thread, since an LLM operates on tokens
rather than characters or meaning, so an edit that preserves meaning can still reshape the
token stream, shift it away from the distribution the model saw in training, or redistribute
attention over key spans. Table~\ref{tab:families} summarizes the eleven families by the
surface they act on and the mechanism by which each can move a model's output, in playground
order; the full per-perturbation catalog is in Appendix~\ref{sec:appendix-catalog}.

\newcolumntype{Y}{>{\raggedright\arraybackslash}X}

\begin{table*}[t]
  \small
  \begin{tabularx}{\textwidth}{@{}l >{\raggedright\arraybackslash}p{2.1cm} Y@{}}
    \toprule
    Family & Acts on & How and why it can affect an LLM \\
    \midrule
    \texttt{control} & the whole input text & Returns the text unchanged: the reference condition. \\
    \midrule
    \texttt{word\_level} & words and tokens & Real-user word noise (typos, dropped stop words, synonyms, splits) resplits familiar words into rare sub-word tokens. \\
    \midrule
    \texttt{character} & characters and glyphs & Character substitutions (homoglyphs, leetspeak, case, zero-width) stay legible but change the code points and tokens. \\
    \midrule
    \texttt{formatting} & layout and markup & Layout edits (wrapping, tabs, fences, list markers) inject structural tokens that fragment phrases, not content. \\
    \midrule
    \texttt{schema} & JSON Schema structure & Schema rewrites (reorder properties, enums, unions) hold the accepted set fixed but change the compiled grammar. \\
    \midrule
    \texttt{normalization} & surface conventions & Equivalent forms (abbreviations, dates, quote style, emoji) mean the same but tokenize differently. \\
    \midrule
    \texttt{padding} & input-text boundaries & Leading/trailing quotes, spaces, or newlines shift every downstream token position; empirically most disruptive. \\
    \midrule
    \texttt{context} & prepended framing & Prepended urgency or a persona frames the whole response while adding no task content. \\
    \midrule
    \texttt{paraphrasing} & wording and structure & Model-backed rewrites (paraphrase, register shift, round-trip translation) hold meaning fixed but move the token distribution. \\
    \midrule
    \texttt{reasoning} & task framing & Model-backed reframings (premise reorder, negation) hold the answer fixed but probe order- and negation-sensitivity. \\
    \midrule
    \texttt{adversarial} & hidden payloads & Prompt-injection probes append instructions a human cannot see (HTML comments, invisible Unicode) that still tokenize. \\
    \bottomrule
  \end{tabularx}
  \caption{The eleven perturbation families, in playground order, by the surface each acts on
  and the token-level mechanism by which it can change a model's output. Each family collects
  several perturbations; Appendix~\ref{sec:appendix-catalog} gives the per-perturbation
  detail.}
  \label{tab:families}
\end{table*}

Two design decisions organize the catalog. First, provenance. The catalog is first-party: every
family above is designed and implemented in this work. Alongside it we offer an optional
\texttt{brittlebench} compatibility tier of ten perturbations that wraps the released
engine of a separate robustness study \citep{romanou2026brittlebench} as a pinned external
dependency, so a harness can reproduce or compare against that study's exact perturbations
through the same interface. The split is about provenance and licensing rather than packaging,
and it is entirely optional: each of the ten behaviors also has a first-party \texttt{native}
implementation, so a user who does not need external comparability never installs the external
engine.

The two implementations also serve as a mutual check, since a native perturbation and
its wrapped counterpart are byte-comparable for a given seed, verifying our reimplementations
against an independent reference rather than merely asserting agreement. Second, the model-backed
rewrites, sixteen in total across the \texttt{paraphrasing}, \texttt{context}, and
\texttt{reasoning} families, call an LLM through an OpenAI-compatible endpoint to produce
transformations a deterministic transform cannot, such as paraphrases. As noted above, these
are best-effort and non-deterministic, and fenced off from the determinism contract.

Benchmark selection is the responsibility of the downstream harness, not of
\texttt{perturber}. An experiment queries \texttt{list\_perturbations()} or
\texttt{GET /perturbations} and filters on the family, the seed-sensitivity, and the declared
parameters. One axis it can filter on is scope. The catalog marks each perturbation
\texttt{in\_scope} when it is a natural, meaning-preserving surface change a real user could
produce, so a look-alike encoding such as \texttt{homoglyph} or \texttt{bidi\_override} is
in scope even though it also underlies spoofing: the text still reads the same and carries no
added instruction.

A handful of entries are marked out of scope precisely because they cross that
line, injecting hidden instructions (\texttt{invisible\_instructions}, \texttt{unicode\_tag\_smuggle})
or otherwise changing the task; they stay callable for defensive testing but a robustness harness
filters them out. We ship no benchmark manifest by design, since a fixed manifest would bake one
experiment's choices into a tool meant to serve many.

\section{Evaluation}
\label{sec:eval}

We evaluate the tool rather than the models it perturbs; the brittleness measurements belong to
the ongoing robustness study we intend to report separately. This section asks a narrower
question: is the tool correct, reproducible, and fast, and does its central claim of
semantics-preservation hold? Every number below is regenerated from a committed, deterministic
harness that reads the installed library. Correctness, transport, and latency run no model
inference; only the fidelity stage calls a judge.

\subsection{Correctness, Parity, and Cost}

The deterministic tier passes the checks its design implies, over a small fixed corpus. All 63
deterministic perturbations are reproducible without exception (identical output over 20 repeats
across five seeds), each behaves correctly on empty input, and each perturbation's declared
\texttt{seed\_sensitive} flag matches its observed behavior; the 16 model-backed perturbations are
excluded as non-deterministic by design.

\begin{figure}[t]
  \centering
  \includegraphics[width=\columnwidth]{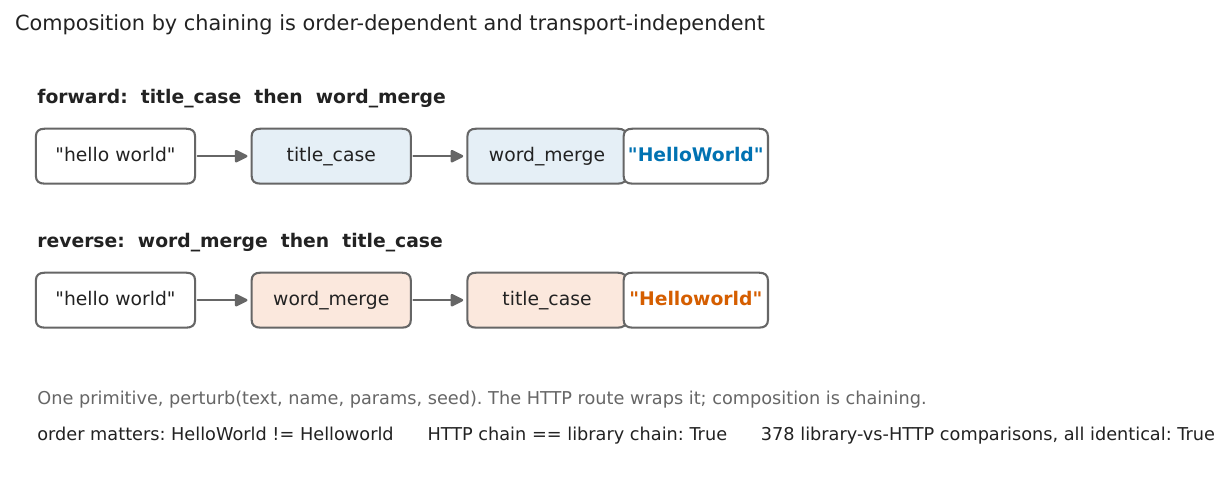}
  \caption{Composition is the caller chaining calls, so it is order-dependent and independent
  of the interface. Chaining \texttt{title\_case} then \texttt{word\_merge} differs from the
  reverse order, and the library chain and the HTTP chain are byte-for-byte identical across
  all 378 comparisons.}
  \label{fig:chaining}
\end{figure}

Interface parity holds byte-for-byte: across all 378
library-versus-\texttt{POST /perturb} comparisons the results were identical, so a chain produces
the same output whichever interface runs it. Composition being caller-side chaining, it is order-dependent
(Figure~\ref{fig:chaining}): \texttt{title\_case} then \texttt{word\_merge} maps \texttt{hello
world} to \texttt{HelloWorld}, and the reverse order differs. The tier is fast, with an overall
median of 23.8 microseconds per call and every family well under 100 microseconds (excluding the
network-bound model-backed rewrites and the optional external engine).

\subsection{Validating Semantics-Preservation}
\label{sec:fidelity}

\begin{figure}[t]
  \centering
  \includegraphics[width=\columnwidth]{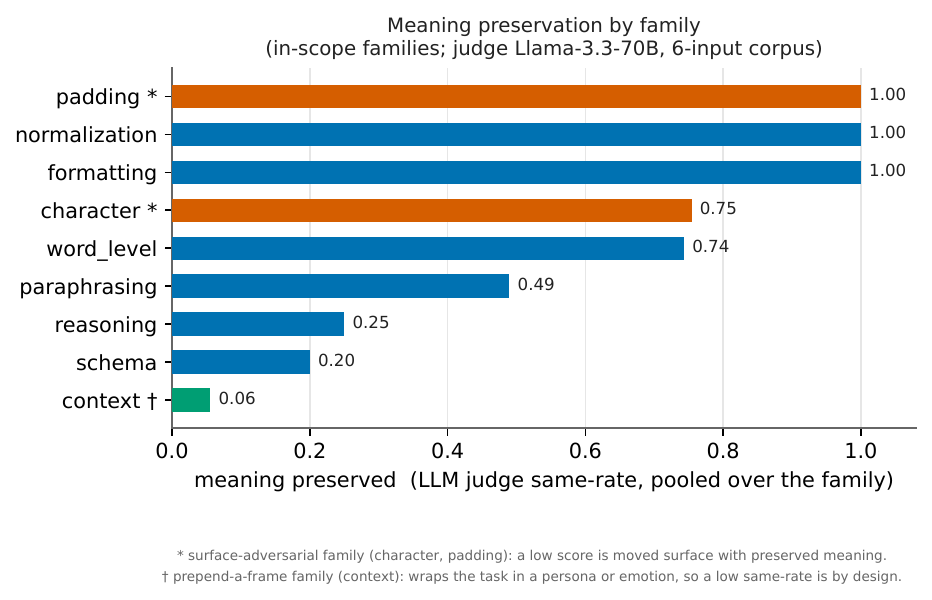}
  \caption{Meaning preserved by family (judge same-meaning rate, pooled per in-scope family).
  Surface-transform families sit at the top. Low bars are low by design: surface-adversarial
  families ($*$) move the surface, and \texttt{context} ($\dagger$) prepends a frame the judge
  reads as a changed message. Schema is low for a different reason (few JSON inputs), discussed in
  the text.}
  \label{fig:family-fidelity}
\end{figure}

\begin{figure}[t]
  \centering
  \includegraphics[width=\columnwidth]{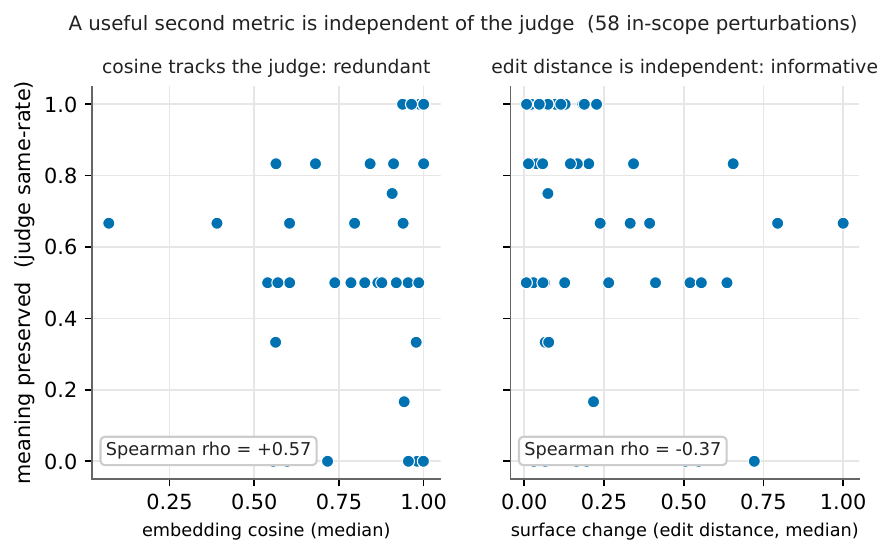}
  \caption{Edit distance keeps the judge honest only if it is independent of it. Embedding cosine
  tracks the judge's same-meaning rate (left, redundant, so it could not tell us the surface moved
  independently of the verdict), whereas edit distance is nearly orthogonal (right, informative), so
  we pair the judge with edit distance.}
  \label{fig:metric-agreement}
\end{figure}

Semantics-preservation is the defining claim of the system, so we measure it along two
deliberately different axes: an LLM judge (Llama-3.3-70B, stronger than and distinct from the model
serving the perturbation tier) for a direct meaning comparison, scoring 313 pairs at temperature
zero with three samples per pair, and normalized edit distance for the magnitude of surface change.
The judge carries the meaning verdict; edit distance keeps it honest, since a same-meaning rate is
only evidence of preservation if the surface actually moved rather than the judge rubber-stamping
near-identical text. Pairing the two rules that out: the in-scope families move the surface
substantially (median edit distance $0.13$ for \texttt{character}, up to $0.52$ for the model-backed
rewrites) yet the judge still reads them as the same question, so surface change is distinct from
meaning change, exactly the property a robustness perturbation must have. Edit distance can play
this role only because it is \emph{independent} of the judge rather than a proxy for it: an embedding
cosine similarity would merely track the judge (Spearman $\rho\!=\!0.57$) and confirm nothing,
whereas surface edit distance is nearly orthogonal to it
($\rho\!=\!-0.37$, Figure~\ref{fig:metric-agreement}).

Preservation is a property of the family rather than of a few favorable points. Grouping the
in-scope perturbations by family (Figure~\ref{fig:family-fidelity}), the surface-transform
families preserve meaning at or near the top of the scale: \texttt{formatting},
\texttt{normalization}, and \texttt{padding} score 1.00, and \texttt{word\_level} 0.74. The low
bars fall into two kinds. The surface-adversarial \texttt{character} family and the
prepend-a-frame \texttt{context} family are low by design: the first moves the surface
deliberately while a reader recovers the same meaning, and the second wraps the task in a persona
or emotional stance the judge reasonably reads as a changed message. The \texttt{schema} family
is low for a different reason we report plainly: it rewrites JSON Schema structure, only one or
two corpus inputs are JSON, and a text-meaning judge reads a structural rewrite as a changed
message, so its score rests on very few pairs and reflects the judge and corpus rather than a
fidelity failure. That the judge scores the by-design families low shows it discriminates: a
negative control confirms this, as the out-of-scope, content-changing perturbations score a mean
same-meaning rate of 0.15, well under a quarter of the in-scope mean of 0.67.
Table~\ref{tab:fidelity-examples} makes the behavior concrete with one worked example per family.

These metrics characterize fidelity rather than guarantee it (see Limitations), and the stage
doubles as a test on the tool itself: it caught two genuine defects in the perturbation layer,
which we corrected.

\input{tables/fidelity_examples.tex}

\subsection{Meaning-Preserving Edits Can Flip Answers}
\label{sec:tokenization}

The fidelity stage shows a perturbation can move the surface while preserving meaning. A
downstream question is what that surface change does to the model, since the same meaning now
reaches it as a different token sequence and can flip the answer. We preview this with a handful
of instances drawn from an ongoing robustness study that applies this library at scale, which we
intend to report in a separate future submission; we include a few flips here only to show the
kind of question the tool is built to ask. The instances are recorded baseline-correct to
perturbed-wrong flips on ARC-Challenge (0-shot, \texttt{acc\_norm}, teacher-forced), where only the
question stem is perturbed and the answer choices and gold answer are held byte-for-byte fixed, so a
flip is a change in which choice the model ranks highest rather than a change in the task.

Every perturbation shown is an edit a real user could make by accident, chosen for exactly that
plausibility, and an independent judge (Llama-3.3-70B, distinct from the scored models) rated each
pair meaning-equivalent, so no flip can be dismissed as a contrived or answer-changing input.
Appendix~\ref{sec:appendix-flips} lists all twelve, three per perturbation, with each perturbed
stem and the model's chosen answer spelled out on both sides; in every row a correct answer turns
wrong.

The edits differ in how much they disturb the surface, from a doubled space or a typo to a
confusable-word swap, and the perturbed token count usually rises with them, yet a flip does not
require it to: a doubled space lengthens the sequence while a confusable-word swap flips the answer
at an unchanged token count. We make no claim about which surface property drives the harm; the
point is narrower. Edits a reader would call harmless, and a judge calls equivalent, still change
the answer, and they do so across visibly different kinds of edit, which is why a robustness study
needs a library of transforms rather than one canonical edit.

The net accuracy change understates this instability, because a perturbation breaks some items while
accidentally repairing others: \texttt{confusable\_words} nets near zero on every model, yet around
fifty answers change under it on each. A study reading only mean accuracy would conclude the edit
does nothing, so the per-item view the tool enables is what exposes the effect.

\section{The Demonstration}
\label{sec:demo}

The live segment exercises the running system through the playground. We enter text and select
a perturbation, and the interface shows the perturbed result with per-item timing. We toggle the
seed to exhibit determinism on the deterministic tier, then run a model-backed rewrite against an
OpenAI-compatible endpoint to show the best-effort tier. We then chain two perturbations to make
order-dependence and transport identity visible, reproducing the worked example of
Section~\ref{sec:eval}. The catalog view doubles as the benchmark-selection interface, listing
each perturbation with its family, seed-sensitivity, parameters, and \texttt{in\_scope} flag, and
it is the same view a harness consumes programmatically through \texttt{GET /perturbations}%
\IfFileExists{figures/shot_playground.png}{~(Figure~\ref{fig:playground}, Appendix)}{}. The
camera-ready version will add a short screencast link.

\section{Related Work}
\label{sec:related}

Controlled, parameterized, reproducible input corruption for evaluating a data-processing
system predates the LLM-robustness framing. In record linkage, for example, GeCo corrupts
personal records under seeded parameter control to stress-test entity-resolution systems
\citep{tran2013geco}. We apply the same seeded-transformation discipline to LLM prompts with the
fidelity goal inverted: GeCo injects error to test how a system copes with dirty data, whereas
\texttt{perturber} preserves meaning while varying the surface to test whether a model's answer
depends on it.

The prompt-robustness literature establishes the phenomenon we build for. A growing body of
work quantifies how much surface-level, meaning-preserving variation moves LLM benchmark
performance: prompt formatting alone induces large accuracy spreads \citep{sclar2024quantifying},
single-prompt evaluation is unreliable enough to warrant reporting across many paraphrases
\citep{mizrahi2024state}, adversarial prompt edits degrade accuracy across tasks
\citep{zhu2023promptrobust}, and broad benchmarks measure brittleness across many models and
perturbations \citep{romanou2026brittlebench}. These are findings papers; \texttt{perturber} complements them as a
reusable transform layer that turns such findings into perturbations a study can reproduce.

Augmentation and behavioral-testing tools are closest to our system in spirit. AugLy
provides multi-modal robustness augmentations \citep{papakipos2022augly}, NLPAug offers a
widely used library of textual augmenters \citep{ma2019nlpaug}, TextAttack frames transformations
around adversarial-attack search \citep{morris2020textattack, jin2020bert} evaluated by benchmarks
such as Adversarial GLUE \citep{wang2021adversarial}, NL-Augmenter collects community
task-sensitive transformations \citep{dhole2021nlaugmenter}, and behavioral suites such as CheckList
\citep{ribeiro2020checklist}, contrast sets \citep{gardner2020evaluating}, and Polyjuice
\citep{wu2021polyjuice} probe model behavior through targeted, meaning-controlled input edits. Our
system differs in five respects: (1) a single-primitive stateless design, so composition is
caller-side chaining; (2) an explicit provenance split between first-party and externally derived
perturbations; (3) a determinism contract, with the seed a first-class call argument and transport
identity; (4) a \texttt{schema} family that perturbs constrained-decoding structure while preserving
the accepted set; and (5) delivery as a live service.

\section{Conclusion}
\label{sec:conclusion}

We presented \texttt{perturber}, a stateless service and Python library that applies exactly one
semantics-preserving perturbation per call. Its single-primitive design makes the library and the
service byte-identical, and its first-party catalog of 68 perturbations lets a downstream harness
select its own benchmark. The wider aim is to decouple \emph{what} a perturbation is from \emph{how} a
study runs it: today each robustness paper reimplements the transforms it needs, so a number
reported under ``typos'' in one study and ``typos'' in another need not describe the same edit. By
naming, scoping, and typing each transform once, and exposing it identically as a library call, a
shell command, and an HTTP route, \texttt{perturber} makes a perturbation a citable object that
robustness studies can share as they share a metric or dataset.

\section*{Acknowledgements}

The design, architecture, and scope of the \texttt{perturber} system were determined by the
authors, who also carried out part of the implementation. AI-based coding and writing assistants
were used to support the remaining implementation and to help edit this paper. The authors
reviewed and verified all code, experiments, and claims, and take full responsibility for the
content.

\section*{Limitations}

Semantics-preservation means different things across families, and we do not claim a single
threshold that holds for all of them; a character-level edit and a paraphrase preserve meaning
in different senses, and the fidelity metrics of Section~\ref{sec:fidelity} report this rather
than paper over it. Those metrics are computed over a small, hand-curated smoke corpus of six
inputs, so they demonstrate that the mechanism works but are not a validated per-perturbation
leaderboard. The model-backed tier is best-effort and non-deterministic, and we exclude it
from every determinism and latency claim. The latency numbers are host-specific, reported
alongside the measurement host and its hardware. Finally, eight perturbations are out of scope
for the project's definition of a robustness perturbation; they remain callable but are flagged
so a harness can exclude them.

\section*{Ethics Statement}

The \texttt{persona\_instruction} perturbation draws its personas from PersonaHub
\citep{ge2024personahub}, released under CC-BY-NC-SA-4.0 for research use only. We
retain this dependency internally so its output stays comparable to the BrittleBench engine's
for a given seed, but for this reason we withhold \texttt{persona\_instruction} from the public
release, whose catalog is therefore the 68 perturbations reported above; we flag the restriction
in the module, the documentation, and the playground so a commercial user is not misled. The catalog also includes a small number of
prompt-injection probes, such as \texttt{invisible\_instructions} and
\texttt{unicode\_tag\_smuggle}. We provide these for defensive evaluation, so a system
can be tested against such inputs; they are flagged out of scope and hidden from the
playground's default view.

\bibliography{references}

\appendix
\section{Reproducing the Evaluation}
\label{sec:appendix-repro}

Every number in Section~\ref{sec:eval} is regenerated by a committed harness. Running
\texttt{python eval/run\_all.py} produces the catalog statistics, the correctness and
determinism matrix, the transport-identity comparison, the native-tier latency table, and the
fidelity metrics, writing each as a machine-readable artifact under \texttt{eval/out/}. The
correctness, transport, and latency stages run no model inference and are safe in
continuous integration; only the fidelity stage calls an embedding model and an LLM judge, and
it reports which were available. The methodology and the map from each claim to the artifact
that backs it are documented in \texttt{demo-eval.md}.

\IfFileExists{figures/shot_playground.png}{%
\begin{figure*}[b]
  \centering
  \includegraphics[width=0.82\textwidth]{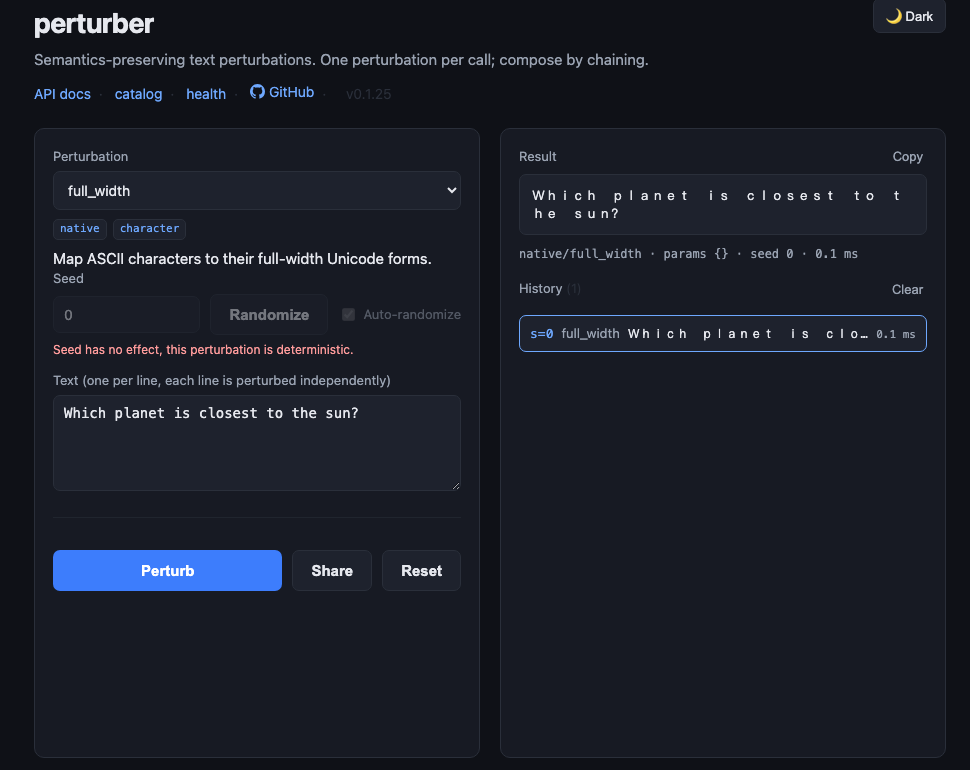}
  \caption{The deployed playground (Section~\ref{sec:demo}). A perturbation is selected from the
  catalog, applied to the input text, and returned with per-item timing; the metadata line
  (\texttt{native/full\_width}~$\cdot$~\texttt{seed 0}~$\cdot$~elapsed) is the same information the
  API returns, and the run history makes determinism visible. This is a thin interface over the
  same primitive a harness calls over HTTP.}
  \label{fig:playground}
\end{figure*}}{}

\section{The Full Catalog}
\label{sec:appendix-catalog}

Tables~\ref{tab:cat-control} through~\ref{tab:cat-reasoning} give the full in-scope catalog by
family, in playground order, with perturbations sorted by name within each family. For each we
state the transformation it applies and the mechanism by which it can change a model's output.
We describe each behavior once; ten are additionally available as a \texttt{brittlebench}
variant that wraps the original study's engine, and the eight out-of-scope perturbations are
omitted here, matching the playground.

\newcommand{\catcaption}[2]{\caption{The \texttt{#1} family: #2}}

\begin{table*}[t]
  \small
  \begin{tabularx}{\textwidth}{@{}l Y Y@{}}
    \toprule
    Perturbation & What it does & Why it can affect an LLM \\
    \midrule
    \texttt{baseline} & Returns the text unchanged. & The reference point: any difference under another perturbation is measured against it, isolating sensitivity from behavior on the clean prompt. \\
    \bottomrule
  \end{tabularx}
  \catcaption{control}{the reference condition.}
  \label{tab:cat-control}
\end{table*}

\begin{table*}[t]
  \small
  \begin{tabularx}{\textwidth}{@{}l Y Y@{}}
    \toprule
    Perturbation & What it does & Why it can affect an LLM \\
    \midrule
    \texttt{apostrophe\_drop} & Drops apostrophes from contractions and possessives (don't becomes dont, it's becomes its). & One of the most common real-word typos; the de-apostrophed form tokenizes differently and can collide with a different word (its versus it's). \\
    \midrule
    \texttt{confusable\_words} & Swaps confusable words for a wrong-but-valid alternative (their/there, to/too, its/it's). & A real-word error substitutes a valid but wrong word, which the model may take at face value rather than flagging as noise. \\
    \midrule
    \texttt{contractions} & Contracts expanded phrases (do not becomes don't). & Contractions and expansions tokenize differently and occur at different training rates, so continuations can differ. \\
    \midrule
    \texttt{drop\_stop\_words} & Removes common stop words, keeping question, negation, and logic words. & Tests reliance on function words for structure; their loss can make parsing ambiguous even with content words intact. \\
    \midrule
    \texttt{number\_format} & Adds thousands separators to large integers (1000 becomes 1,000). & Changes how digits tokenize, a known source of arithmetic and quantity-reasoning errors. \\
    \midrule
    \texttt{punctuation\_drop} & Drops terminal punctuation and lowercases sentence starts. & Casual, unpunctuated input removes the boundary cues the model uses to segment thoughts. \\
    \midrule
    \texttt{punctuation\_spaces} & Inserts spaces around punctuation. & Detaching punctuation changes tokenization at clause boundaries, blurring where one thought ends and the next begins. \\
    \midrule
    \texttt{realistic\_typos} & Models how people actually mistype: keyboard slips, common misspellings, transpositions, doubled or dropped keys, vowel confusions, missed-space merges, phonetic slips, and case errors. & Probes robustness to the realistic noisy input real users produce, not just uniform random edits. \\
    \midrule
    \texttt{repeated\_chars} & Stretches words for emphasis (so becomes sooo). & Stretched spellings are rare out-of-distribution tokens, common in casual chat but absent from clean benchmark text. \\
    \midrule
    \texttt{sequence\_spaces} & Inserts random-length space runs between tokens. & Extra whitespace produces unusual token sequences and can dilute attention, a surface change with no meaning change. \\
    \midrule
    \texttt{spelling\_uk} & Converts American spellings to British (color becomes colour). & Regional spellings differ in training frequency and tokenization, a benign shift that can still move the output. \\
    \midrule
    \texttt{synonym} & Replaces content words with WordNet synonyms, chosen by word sense in context. & Word choice strongly drives the next-token distribution; a meaning-preserving swap can steer the response. \\
    \midrule
    \texttt{typos} & Introduces character-level typos into words. & Typos split common words into unusual sub-word tokens rarely seen intact, degrading the model's grasp of the term. \\
    \midrule
    \texttt{word\_merge} & Removes the space between two adjacent words. & The merged form tokenizes very differently, so a familiar phrase becomes an unfamiliar token run. \\
    \midrule
    \texttt{word\_split} & Inserts a space inside a word. & The fragments tokenize as unrelated pieces, so the model may fail to recover the intended word. \\
    \bottomrule
  \end{tabularx}
  \catcaption{word\_level}{word- and token-level edits.}
  \label{tab:cat-word}
\end{table*}

\begin{table*}[t]
  \small
  \begin{tabularx}{\textwidth}{@{}l Y Y@{}}
    \toprule
    Perturbation & What it does & Why it can affect an LLM \\
    \midrule
    \texttt{bidi\_override} & Wraps the text in a Unicode right-to-left override, so it renders reversed in many viewers. & Probes robustness to bidirectional-control-character spoofing, where displayed and encoded order diverge. \\
    \midrule
    \texttt{diacritics\_add} & Adds combining accents to vowels. & Accented forms are rarer tokens, so familiar words become unfamiliar ones. \\
    \midrule
    \texttt{diacritics\_strip} & Removes all diacritics (caf\'e becomes cafe). & Where accents are meaning-bearing, stripping them changes both tokens and, at the margin, sense. \\
    \midrule
    \texttt{disemvowel} & Drops a subset of vowels (hello becomes hll). & Humans still read it; the model must reconstruct heavily abbreviated, out-of-distribution tokens. \\
    \midrule
    \texttt{full\_width} & Maps ASCII to full-width Unicode forms. & Visually similar, tokenized completely differently, and rare in training. \\
    \midrule
    \texttt{homoglyph} & Replaces letters with visually identical Unicode confusables (Latin a becomes Cyrillic a). & Humans see the same word; the model sees entirely different code points and tokens. This also underlies spoofing attacks. \\
    \midrule
    \texttt{leetspeak} & Substitutes lookalike digits and symbols (e becomes 3, o becomes 0). & Breaks words into atypical tokens while a person still reads the original. \\
    \midrule
    \texttt{random\_case} & Flips the case of a subset of letters. & Casing changes tokenization (Hello versus hELLo); models are case-sensitive despite identical meaning. \\
    \midrule
    \texttt{title\_case} & Title-cases the whole text. & Case shifts alter token boundaries throughout and can be read as emphasis or heading style. \\
    \midrule
    \texttt{upper\_case} & Uppercases the whole text. & Case shifts alter token boundaries throughout and can be read as emphasis or shouting. \\
    \midrule
    \texttt{zero\_width} & Inserts invisible zero-width spaces between characters. & Nothing is visible to a reader, yet the model sees extra tokens splitting words, a pure token-versus-text mismatch. \\
    \bottomrule
  \end{tabularx}
  \catcaption{character}{character- and glyph-level edits.}
  \label{tab:cat-character}
\end{table*}

\begin{table*}[t]
  \small
  \begin{tabularx}{\textwidth}{@{}l Y Y@{}}
    \toprule
    Perturbation & What it does & Why it can affect an LLM \\
    \midrule
    \texttt{bullet\_reformat} & Converts numbered list markers to dash bullets. & List-style drift changes structural tokens without changing the items, testing sensitivity to formatting. \\
    \midrule
    \texttt{extra\_whitespace} & Doubles the space at a subset of word boundaries. & Perturbs token positions without touching meaning, like \texttt{sequence\_spaces}. \\
    \midrule
    \texttt{line\_wrap} & Hard-wraps the text at a fixed column width. & Hard line breaks mid-sentence, as from an email or terminal paste, inject newline tokens that fragment phrases. \\
    \midrule
    \texttt{markdown\_wrap} & Wraps the text in a Markdown code fence. & Fencing can flip the model between treating content as prose versus code or verbatim. \\
    \midrule
    \texttt{strip\_markdown} & Removes Markdown markup, leaving the plain text. & Users often paste rendered text; losing emphasis and structure markers can drop cues the model relied on. \\
    \midrule
    \texttt{tabs\_for\_spaces} & Replaces runs of spaces with tabs. & Whitespace style changes tokenization and can matter in code or structured prompts. \\
    \bottomrule
  \end{tabularx}
  \catcaption{formatting}{layout and markup changes.}
  \label{tab:cat-formatting}
\end{table*}

\begin{table*}[t]
  \small
  \begin{tabularx}{\textwidth}{@{}l Y Y@{}}
    \toprule
    Perturbation & What it does & Why it can affect an LLM \\
    \midrule
    \texttt{add\_redundant\_constraints} & Adds tautological keywords (\texttt{minItems} or \texttt{minProperties} of 0) to a JSON Schema. & Never narrows the accepted set; probes sensitivity to grammar text that carries no new information. \\
    \midrule
    \texttt{expand\_type\_shorthand} & Rewrites a \texttt{type} list as an equivalent \texttt{anyOf} of single-type subschemas. & Same accepted set, different compiled grammar, testing whether the decoding path itself moves the output. \\
    \midrule
    \texttt{inline\_defs} & Inlines local \texttt{\$ref} definitions and drops the unused \texttt{\$defs}. & An equivalent schema with a different structure; the compiled grammar differs while the language is identical. \\
    \midrule
    \texttt{reorder\_enum} & Permutes the order of every \texttt{enum} value list. & The decoder explores enum alternatives in order, which can bias which categorical value the model settles on. \\
    \midrule
    \texttt{reorder\_properties} & Permutes the key order of every \texttt{properties} object (and \texttt{required}). & Constrained decoders emit keys in schema order, so this changes the generation order (what each field is conditioned on, e.g.\ reasoning before answer) with the accepted document set unchanged. \\
    \midrule
    \texttt{reorder\_union} & Permutes the branch order of every \texttt{anyOf}, \texttt{oneOf}, and \texttt{allOf}. & Branch order affects which union alternative the constrained decoder explores first. \\
    \bottomrule
  \end{tabularx}
  \catcaption{schema}{meaning-preserving JSON Schema rewrites for constrained decoding.}
  \label{tab:cat-schema}
\end{table*}

\begin{table*}[t]
  \small
  \begin{tabularx}{\textwidth}{@{}l Y Y@{}}
    \toprule
    Perturbation & What it does & Why it can affect an LLM \\
    \midrule
    \texttt{abbreviations} & Expands common abbreviations (e.g.\ becomes for example). & Expanded and abbreviated forms mean the same but tokenize differently and cue different registers. \\
    \midrule
    \texttt{date\_unit} & Reformats ISO dates and spells out units after numbers. & Equivalent date and unit surface forms are a common source of extraction and normalization errors. \\
    \midrule
    \texttt{emoji} & Converts text emoticons to emoji. & An emoticon and its emoji encode the same sentiment but tokenize differently and occur at different training rates. \\
    \midrule
    \texttt{emoticon} & Converts emoji to text emoticons. & The inverse of \texttt{emoji}: the same sentiment tokenizes differently across the two forms. \\
    \midrule
    \texttt{quote\_style} & Swaps straight quotes for typographic (curly) quotes. & Curly and straight quotes are different code points and tokens, so quoting style alone can move the output. \\
    \midrule
    \texttt{smart\_to\_ascii} & Folds typographic quotes, dashes, and ellipses to plain ASCII. & The inverse of \texttt{quote\_style}: what a word processor's smart punctuation becomes on paste into a plain-text field, changing tokens. \\
    \bottomrule
  \end{tabularx}
  \catcaption{normalization}{surface-form normalization.}
  \label{tab:cat-normalization}
\end{table*}

\begin{table*}[t]
  \small
  \begin{tabularx}{\textwidth}{@{}l Y Y@{}}
    \toprule
    Perturbation & What it does & Why it can affect an LLM \\
    \midrule
    \texttt{new\_lines} & Prepends and appends $N$ newline characters. & Leading and trailing padding shifts every downstream token position and can read as a delimiter, among the most impactful types in the study. \\
    \midrule
    \texttt{quotes} & Prepends and appends $N$ quote characters. & Leading and trailing padding shifts every downstream token position and can read as a delimiter. \\
    \midrule
    \texttt{spaces} & Prepends and appends $N$ space characters. & Leading and trailing padding shifts every downstream token position and can read as a delimiter. \\
    \bottomrule
  \end{tabularx}
  \catcaption{padding}{leading and trailing padding.}
  \label{tab:cat-padding}
\end{table*}

\begin{table*}[t]
  \small
  \begin{tabularx}{\textwidth}{@{}l Y Y@{}}
    \toprule
    Perturbation & What it does & Why it can affect an LLM \\
    \midrule
    \texttt{emotion\_prompt} & Prepends an emotional-urgency phrase (model-backed). & EmotionPrompt-style affective framing shifts model performance despite adding no task content. \\
    \midrule
    \texttt{persona\_instruction} & Prepends a persona directive, with the seed selecting the persona. & Injecting a role frames the whole response, a well-documented lever on model behavior and accuracy. \\
    \bottomrule
  \end{tabularx}
  \catcaption{context}{framing prepended before the text.}
  \label{tab:cat-context}
\end{table*}

\begin{table*}[t]
  \small
  \begin{tabularx}{\textwidth}{@{}l Y Y@{}}
    \toprule
    Perturbation & What it does & Why it can affect an LLM \\
    \midrule
    \texttt{lexical\_paraphrasing} & Rewords with different vocabulary, keeping structure and meaning (model-backed). & Word choice shifts the token distribution; the same question phrased differently can get a different answer. \\
    \midrule
    \texttt{persona\_paraphrase} & Rewrites in the voice of a plausible persona (model-backed). & Voice and style are a training sub-distribution lever; unlike \texttt{persona\_instruction} it is model-generated, so no dataset license applies. \\
    \midrule
    \texttt{register\_shift} & Rewrites into a chosen register via the \texttt{target\_register} parameter (formal, casual, academic, legal, or child; model-backed). & Register correlates with different training sub-distributions, so register alone can change the response; the parameter sweeps that axis. \\
    \midrule
    \texttt{rule\_free\_paraphrasing} & Free-form paraphrase preserving meaning (model-backed). & Measures overall sensitivity to phrasing. It is confounded, since paraphrasing can also improve accuracy. \\
    \midrule
    \texttt{simplification} & Rewrites in simpler, plainer language (model-backed). & Reading-level shifts change vocabulary and structure, testing whether the answer depends on phrasing sophistication. \\
    \midrule
    \texttt{syntactic\_paraphrasing} & Restructures grammar, such as clause order or active versus passive, keeping wording and meaning (model-backed). & Sentence structure affects how attention distributes over the prompt. \\
    \midrule
    \texttt{translate\_roundtrip} & Paraphrases by round-tripping through a pivot language via the \texttt{language} parameter (model-backed). & Round-trip translation yields idiomatic rewrites; a more distant pivot drifts more, so the parameter sweeps paraphrase strength. \\
    \midrule
    \texttt{typo\_paraphrase} & Rewrites with natural, model-generated human typos (model-backed). & Contrasts with the deterministic \texttt{realistic\_typos}: model-generated noise and synthetic edits may degrade models differently. \\
    \midrule
    \texttt{verbose\_expansion} & Expands into wordier, more roundabout phrasing (model-backed). & Padding and filler dilute the salient tokens and lengthen the context the model must attend over. \\
    \bottomrule
  \end{tabularx}
  \catcaption{paraphrasing}{model-backed meaning-preserving rewrites.}
  \label{tab:cat-paraphrasing}
\end{table*}

\begin{table*}[t]
  \small
  \begin{tabularx}{\textwidth}{@{}l Y Y@{}}
    \toprule
    Perturbation & What it does & Why it can affect an LLM \\
    \midrule
    \texttt{negation\_paraphrase} & Restates via meaning-preserving negations or double negatives (model-backed). & Models are brittle to negation; a meaning-preserving negation rewrite probes that directly. \\
    \midrule
    \texttt{reasoning\_restructure} & Reorders or inverts how the task is framed, keeping the task and answer (model-backed). & Premise and information order measurably affect reasoning; this isolates order from content. \\
    \bottomrule
  \end{tabularx}
  \catcaption{reasoning}{reframings that hold the task fixed.}
  \label{tab:cat-reasoning}
\end{table*}

\section{All Twelve Answer Flips}
\label{sec:appendix-flips}

Table~\ref{tab:tokenization-flips-appendix} lists all twelve answer flips previewed in
Section~\ref{sec:tokenization}, three per perturbation, with each perturbed stem and the model's
chosen answer spelled out on both sides, so a reader can see that every case is a plausible edit
and a real wrong answer.

\input{tables/tokenization_flips_appendix.tex}

\end{document}

%% file: tables/fidelity_examples.tex
\begin{table*}[t]
\centering
\small
\begin{tabularx}{\textwidth}{@{}l l Y Y c@{}}
\toprule
Family & Perturbation & Input & Perturbed & Judge \\
\midrule
character & \texttt{title\_case} & What is the capital of France\ldots{} & What Is The Capital Of France\ldots{} & same \\
\midrule
context & \texttt{emotion\_prompt} & What is the capital of France\ldots{} & I'm really eager to know the \ldots{} & same \\
\midrule
formatting & \texttt{extra\_whitespace} & The quick brown fox jumps ove\ldots{} & The quick  brown fox  jumps o\ldots{} & same \\
\midrule
normalization & \texttt{quote\_style} & \{"type":"object","properties"\ldots{} & \{``type'':``object'',``properties''\ldots{} & same \\
\midrule
padding & \texttt{quotes} & The quick brown fox jumps ove\ldots{} & """"" The quick brown fox jum\ldots{} & same \\
\midrule
paraphrasing & \texttt{rule\_free\_paraphrasing} & The quick brown fox jumps ove\ldots{} & The swift brown fox leaps ove\ldots{} & same \\
\midrule
reasoning & \texttt{reasoning\_restructure} & What is the capital of France\ldots{} & Why did the capital of France\ldots{} & same \\
\midrule
schema & \texttt{reorder\_union} & \{"type":"object","properties"\ldots{} & \{"type":"object","properties"\ldots{} & same \\
\midrule
word\_level & \texttt{word\_split} & Their going to the store to b\ldots{} & Thei r going to the stor e to\ldots{} & same \\
\bottomrule
\end{tabularx}
\caption{One worked in-scope example per family (input $\rightarrow$ perturbed, judge Llama-3.3-70B's
same-meaning verdict), each its family's median-surface meaning-preserving pair.
Strings are clipped for display.}
\label{tab:fidelity-examples}
\end{table*}

%% file: tables/tokenization_flips_appendix.tex
\begin{table*}[t]
\centering
\small
\begin{tabularx}{\textwidth}{@{}l Y Y@{}}
\toprule
Perturbation & Stem (original $\rightarrow$ perturbed) & Answer (original $\rightarrow$ perturbed) \\
\midrule
\texttt{realistic\_typos} & What best describes skin? $\rightarrow$ Whaft best describeSS skin? & B (flexible) $\rightarrow$ C (brittle) \\
 & Which is a chemical change? $\rightarrow$ hWahc is a chemical change? & A (frying an egg) $\rightarrow$ D (crushing a piece of bread) \\
 & Which describes only a physical change? $\rightarrow$ Whach deescribes oonly a physical change? & D (squeezing the juice from fruit) $\rightarrow$ A (baking a cake) \\
\midrule
\texttt{repeated\_chars} & Which is a renewable resource? $\rightarrow$ Which isss aaaa renewable resource? & C (trees) $\rightarrow$ D (aluminum) \\
 & How are sedimentary rocks made? $\rightarrow$ How areeee sedimentary rocksss made? & B (Materials are pressed together.) $\rightarrow$ C (Chemical reactions change minerals.) \\
 & Which describes a chemical change? $\rightarrow$ Whichhhh describes a chemical change???? & D (acid in the stomach digesting food) $\rightarrow$ C (scissors cutting paper into pieces) \\
\midrule
\texttt{extra\_whitespace} & Which of these objects is opaque? $\rightarrow$ Which\textvisiblespace{}\textvisiblespace{}of these\textvisiblespace{}\textvisiblespace{}objects is opaque? & A (A brick wall) $\rightarrow$ D (Stained glass windows) \\
 & Which statement concerning metals is correct? $\rightarrow$ Which\textvisiblespace{}\textvisiblespace{}statement concerning metals is\textvisiblespace{}\textvisiblespace{}correct? & B (Metals conduct electricity.) $\rightarrow$ D (Metals have a low melting point.) \\
 & Which statement best describes a property of water? $\rightarrow$ Which statement best\textvisiblespace{}\textvisiblespace{}describes a property\textvisiblespace{}\textvisiblespace{}of water? & B (Water changes to a gas as it boils.) $\rightarrow$ A (Liquid water holds its shape.) \\
\midrule
\texttt{confusable\_words} & Which of the following is NOT a form of precipitation? $\rightarrow$ Which off the following is NOT a form of precipitation? & B (Wind) $\rightarrow$ D (Snow) \\
 & Which of the following is most likely a learned behavior? $\rightarrow$ Which off the following is most likely a learned behavior? & A (using a fork to eat) $\rightarrow$ B (jumping at a loud noise) \\
 & What is the main function of photosynthetic cells within a p... $\rightarrow$ What is the main function off photosynthetic cells within a ... & C (to convert energy from sunlight into food energy) $\rightarrow$ D (to allow the passage of carbon dioxide into the plant) \\
\bottomrule
\end{tabularx}
\caption{All twelve answer flips previewed in Section~\ref{sec:tokenization}, three per perturbation (ARC-Challenge, 0-shot, \texttt{acc\_norm}, teacher-forced; only the stem perturbed, choices and gold held fixed). \emph{Answer} gives the model's top choice and its text on each side (gold $\rightarrow$ wrong); an independent judge (Llama-3.3-70B) rated every pair meaning-equivalent. Two numbers of different kinds are quoted here and should not be read together: the study-wide mean $\Delta$acc (pp, over 1{,}172 items) per perturbation is \texttt{realistic\_typos} $-2.30$, \texttt{repeated\_chars} $-1.88$, \texttt{extra\_whitespace} $-0.34$, \texttt{confusable\_words} $-0.85$, while any single token count below is one illustrative example, not evidence. Read that way, the token counts only illustrate variety of surface disturbance: the perturbed count usually rises (\texttt{extra\_whitespace} $7\rightarrow9$), but a flip does not need it to (\texttt{confusable\_words} flips at an unchanged $20\rightarrow20$). Doubled spaces are marked \textvisiblespace\textvisiblespace{} so the \texttt{extra\_whitespace} edit is visible in print. Strings are clipped for display.}
\label{tab:tokenization-flips-appendix}
\end{table*}